\documentclass{article}
\usepackage{spconf,amsmath,amssymb,amsfonts,graphicx}
\usepackage{algorithmic}
\usepackage{textcomp}
\usepackage{booktabs}
\usepackage{pifont}
\usepackage{xcolor}
\usepackage{cite}
\usepackage{hyperref}
\usepackage{siunitx}
\usepackage[utf8]{inputenc}
\usepackage{newunicodechar}
\newunicodechar{①}{1}
\newunicodechar{②}{2}
\newunicodechar{③}{3}
\newunicodechar{④}{4}

\usepackage{adjustbox}

\title{FLASH: Efficient Impact Fall Detection with Unified Hypergraph State-Space Model}

\name{Tresor Y. Koffi$^{1}$, Youssef Mourchid$^{1}$, Yohan Dupuis$^{1}$}
\address{$^{1}$CESI, CESI LINEACT, France\\
\{ytkoffi, ymourchid, ydupuis\}@cesi.fr}

\begin{document}

\maketitle

\begin{abstract}
Falls represent a critical public health challenge, and accurate detection of the impact moment when an individual hits the ground is crucial for timely intervention. Existing skeleton-based methods rely on graph neural networks modeling only pairwise joint connections, failing to capture multi-joint coordination characteristic of fall impacts, while transformer-based temporal models suffer from quadratic complexity limiting real-time deployment. We propose FLASH, a novel framework integrating single-matrix hypergraph representations with Mamba's selective state-space models through adaptive feedback mechanisms for efficient impact detection. Our approach constructs biomechanically-grounded hyperedges to model functional joint coordination while leveraging Mamba's linear-time complexity to capture temporal dynamics. Experiments on UP-Fall and UMAFall datasets demonstrate that FLASH achieves state-of-the-art accuracy with real-time inference capability and strong zero-shot cross-dataset generalization, while significantly reducing computational cost compared to dual-representation and transformer-based methods. The model provides interpretable feedback through learned attention patterns aligned with biomechanical principles. Code is available at \url{https://github.com/Tresor-Koffi/FLASH-Impact-Fall-Detection}.

\end{abstract}


\section{Introduction}
\label{sec:intro}

Falls rank as the second leading cause of unintentional injury-related deaths worldwide \cite{kannus1999fall}, making robust fall detection systems increasingly critical \cite{martinez2019upfall} \cite{koffi2023machine}. Effective intervention requires not only detecting that a fall has occurred but also identifying the precise moment of ground impact the instant when the body hits the ground. This temporal precision is essential for assessing injury risk and triggering timely responses. Moreover, accurate impact detection must distinguish actual falls from visually similar activities such as stumbling, bending to pick up objects, or sitting down quickly, which can otherwise lead to false alarms.
Skeleton-based fall detection has gained prominence due to advances in pose estimation that enable reliable extraction of 3D joint trajectories from video \cite{chen2020fall}. Graph neural networks (GNNs) model skeletal structure by representing joints as nodes and bones as edges \cite{yan2018spatial}, with spatial-temporal graph convolutional networks (ST-GCNs) achieving strong performance by jointly modeling spatial topology and temporal dynamics \cite{yan2018spatial,keskes2021vision}.
However, GNNs are limited by their reliance on pairwise relationships, decomposing multi-joint coordination into independent two-node interactions. This assumption breaks down during fall impacts, where forces propagate through coordinated joint groups. Hypergraphs address this by allowing hyperedges to connect arbitrary joint sets \cite{feng2019hypergraph}, but existing methods employ dual-matrix representations or adaptive hyperedge learning that increase computational overhead.
Temporal modeling poses a second challenge. Transformers suffer from quadratic complexity, while recurrent architectures struggle with long-range dependencies\cite{xin2023transformer}. Mamba, a selective state-space model, offers linear-time complexity and input-dependent memory retention ideal for real-time processing \cite{gu2023mamba}.
To address these challenges, we propose FLASH, a novel architecture for efficient impact detection. Our contributions are:
\begin{enumerate}

\item We used a single-matrix hypergraph representation
for human skeleton to detect impact, which encoded
multi-joint coordination while avoiding dual-incidence
redundancy.

\item We integrate hypergraph spatial modeling with Mamba's selective state-space architecture for efficient temporal processing , achieving 71.3\% reduction in FLOPs and 93.9\% faster inference compared to dual-representation methods, as validated through ablation studies.

\item We provide interpretable feedback through learned joint attention patterns, demonstrating that our model automatically focuses on biomechanically relevant joints for impact detection.


\item We validated our approach on UP-Fall \cite{martinez2019upfall} and UMAFall \cite{casilari2017umafall} datasets, achieving state-of-the-art performance with strong cross-dataset generalization and real-time inference.

\end{enumerate}


The remainder of this paper is organized as follows: Section~\ref{sec:related} reviews related work. Section~\ref{sec:method} details our methodology. Section~\ref{sec:experiments} presents experimental results, and Section~\ref{sec:conclusion} concludes the paper.
\section{Related Work}
\label{sec:related}
\subsection{Graph-Based Skeleton Action Recognition}

Graph neural networks have become the dominant approach for skeleton-based action recognition, representing joints as nodes and bones as edges \cite{yan2018spatial}. Spatial-Temporal Graph Convolutional Networks (ST-GCNs) jointly model spatial topology and temporal dynamics, achieving strong performance in both action recognition and fall detection \cite{yan2018spatial, keskes2021vision}. Subsequent improvements include adaptive graph structures, attention mechanisms, and multi-scale temporal modeling \cite{yan2023skeleton}.
However, GCNs are fundamentally limited to pairwise interactions \cite{koffi2025impact}. This assumption breaks down during impact events, where forces propagate through coordinated joint groups rather than independent two-node connections. Additionally, fixed bone-based adjacency matrices cannot capture functional groupings that emerge during falls. These limitations motivate hypergraph representations that model multi-joint coordination directly.

\subsection{Hypergraph Neural Networks}

Hypergraphs generalize graphs by allowing hyperedges to connect arbitrary numbers of vertices, enabling higher-order relationship modeling \cite{feng2019hypergraph}. This foundational work by Feng et al. provides an accessible introduction to hypergraph neural networks. For skeleton-based recognition, Hao et al. \cite{hao2021hypergraph} proposed Hyper-GNN to capture multiscale spatiotemporal patterns. Recent advances include AutoregAd-HGformer \cite{ray2025autoregressive} with adaptive hyperedge generation and HyperMV \cite{gao2024hypergraph} for multi-view recognition. While effective for general action recognition, these approaches share limitations for real-time impact detection: dual-representation architectures processing both first-order and second-order incidence matrices, computationally expensive adaptive structure learning, and optimization for diverse actions rather than specialized impact detection. Our single-matrix hypergraph with fixed biomechanical hyperedges addresses these limitations.
\subsection{Temporal Modeling and State-Space Models}
Temporal modeling for fall detection has evolved through three paradigms. RNNs and LSTMs capture sequential dependencies \cite{guan2017ensembles} but suffer from vanishing gradients and limited long-range modeling. Transformers overcome these issues through self-attention \cite{xin2023transformer}, but their quadratic complexity  prohibits real-time deployment on edge devices \cite{guo2025tcformer}.
State-space models (SSMs), particularly Mamba \cite{gu2023mamba} \cite{koffi2025distillh}, offer a compelling alternative with linear complexity and selective information. Unlike transformers that distribute attention globally, Mamba's input-dependent parameters enable focused processing of abrupt kinematic transitions characteristic of impact moments. This efficiency-accuracy balance makes SSMs ideal for real-time fall detection, motivating our integration of Mamba with hypergraph spatial modeling.
\section{Methodology}
\label{sec:method}

\subsection{Problem Formulation}

Given a skeletal motion sequence $\mathbf{S} = \{s_1, s_2, \ldots, s_T\}$ where each $s_t$ represents 3D joint positions at time $t$, we aim to identify the specific frame $s_k$ where impact occurs. Each frame at time step $t$ contains $J$ joints in 3D space: $s_t = \{j_1^t, j_2^t, \ldots, j_J^t\}$ where $j_i^t \in \mathbb{R}^3$ represents the $(x, y, z)$ coordinates of joint $i$ at time $t$.
This formulation differs significantly from traditional fall detection, which typically provides a binary classification for the entire sequence. Our approach addresses the more challenging problem of identifying the precise moment within a fall sequence when the body makes contact with the ground critical information for assessing fall severity and coordinating medical response.

\subsection{Single-Matrix Hypergraph Construction}
\label{sec:hypergraph}

We represent the human skeleton as a hypergraph $\mathcal{G} = (\mathcal{V}, 
\mathcal{E})$, where vertices $\mathcal{V}$ correspond to $J = 33$ skeletal 
joints and hyperedges $\mathcal{E}$ group joints with coordinated motion during 
impacts.Following biomechanical principles \cite{murakami2024finite, 
yang2020effect}, we construct $E = 6$ anatomical hyperedges grouping torso and 
lower extremity joints to capture force transmission during impact. The 
incidence matrix $\mathbf{H} \in \mathbb{R}^{J \times E}$ encodes membership 
($H_{i,j} = 1$ if joint $i$ belongs to hyperedge $j$, else $0$), normalized as:

\begin{equation}
\mathbf{H}_{\text{norm}} = \mathbf{D}_v^{-1/2} \mathbf{H} \mathbf{D}_e^{-1} 
\mathbf{H}^T \mathbf{D}_v^{-1/2}
\label{eq:hypergraph_norm}
\end{equation}

where $\mathbf{D}_v$ and $\mathbf{D}_e$ are vertex and hyperedge degree matrices 
\cite{gao2024hypergraph}, preventing high-degree joints from dominating feature aggregation.

\subsection{Hypergraph Convolution with Single Matrix}
\label{sec:hgcn}

Given input features $\mathbf{X} \in \mathbb{R}^{T \times J \times C}$ where $T$ is sequence length and $C = 3$ corresponds to the $(x, y, z)$ joint coordinates, we apply hypergraph convolution using only $\mathbf{H}_{\text{norm}}$:

\begin{equation}
\mathbf{X}^{(l+1)} = \sigma\left(\mathbf{H}_{\text{norm}} \mathbf{X}^{(l)} \mathbf{W}^{(l)}\right)
\label{eq:hgcn}
\end{equation}

where $\sigma$ is ReLU activation and $\mathbf{W}^{(l)}$ are learnable parameters. We apply two layers of hypergraph convolution to extract spatial features that capture multi-joint coordination patterns. The output $\mathbf{F}_{\text{hyper}} \in \mathbb{R}^{T \times J \times C'}$ preserves the joint-wise structure while encoding higher-order relationships, where $C' = 128$ is the hidden feature dimension.

\subsection{Mamba Integration for Temporal Modeling}
\label{sec:mamba}

While hypergraphs capture multi-joint spatial relationships, modeling temporal 
dynamics requires efficient long-range sequence modeling. We integrate Mamba's 
selective state-space models, which provide linear-time complexity versus 
transformers' quadratic complexity, crucial for real-time deployment.

\subsubsection{Selective State-Space Models}
Mamba extends standard SSMs with input-dependent parameters for selective 
information propagation. The continuous and discrete SSMs are defined as:

\begin{align}
\frac{d\mathbf{h}(t)}{dt} &= \mathbf{A}(t)\mathbf{h}(t) + \mathbf{B}(t)\mathbf{x}(t), \quad
\mathbf{y}(t) = \mathbf{C}(t)\mathbf{h}(t) + \mathbf{D}\mathbf{x}(t)
\end{align}
\begin{align}
\mathbf{h}_k &= \tilde{\mathbf{A}}_k\mathbf{h}_{k-1} + \tilde{\mathbf{B}}_k\mathbf{x}_k, \quad
\mathbf{y}_k = \mathbf{C}_k\mathbf{h}_k + \mathbf{D}\mathbf{x}_k
\end{align}

where $\tilde{\mathbf{A}}_k = \exp(\Delta\mathbf{A}_k)$, $\tilde{\mathbf{B}}_k = 
(\Delta\mathbf{A}_k)^{-1}(\exp(\Delta\mathbf{A}_k) - \mathbf{I})(\Delta\mathbf{B}_k)$, 
and $\mathbf{A}$, $\mathbf{B}$, $\mathbf{C}$ are input-dependent through learnable projections.

\subsubsection{Hypergraph-Mamba Integration}
Hypergraph features $\mathbf{F}_{\text{hyper}}$ are reshaped to 
$\mathbf{X}_{\text{seq}} \in \mathbb{R}^{T \times (J \cdot C')}$, processed 
through a MambaBlock, and reshaped back to $\mathbf{F}_{\text{mamba}} \in 
\mathbb{R}^{T \times J \times C''}$. This design maintains Mamba's efficiency 
while benefiting from hypergraph-encoded spatial coordination patterns.

\subsection{Multi-Scale Temporal Processing}
\label{sec:temporal}

Impact events exhibit characteristic temporal patterns at different scales. To capture these dynamics, we apply temporal convolutions to $\mathbf{F}_{\text{mamba}}$:

\begin{align}
\mathbf{Z}_1 &= \sigma(\text{Conv2D}(\mathbf{F}_{\text{mamba}}, (9, 1))) \\
\mathbf{Z}_2 &= \sigma(\text{Conv2D}(\mathbf{F}_{\text{mamba}}, (15, 1))) \\
\mathbf{Z}_3 &= \sigma(\text{Conv2D}(\mathbf{F}_{\text{mamba}}, (20, 1)))
\label{eq:multiscale}
\end{align}

where kernel sizes $(9,1)$, $(15,1)$, $(20,1)$ capture fine, medium, and coarse temporal patterns respectively, while preserving joint dimension. The outputs are concatenated:

\begin{equation}
\mathbf{Z}_{\text{multi}} = \text{Concat}(\mathbf{Z}_1, \mathbf{Z}_2, \mathbf{Z}_3)
\label{eq:concat}
\end{equation}

Finally, fully connected layers produce frame-wise binary classifications:

\begin{equation}
\hat{y}_t = \text{FC}(\mathbf{z}_t), \quad t = 1, \ldots, T
\label{eq:classification}
\end{equation}

where $\hat{y}_t \in \{0,1\}$ indicates impact at frame $t$.

\begin{figure*}[t]
\centering
\includegraphics[width=1\linewidth]{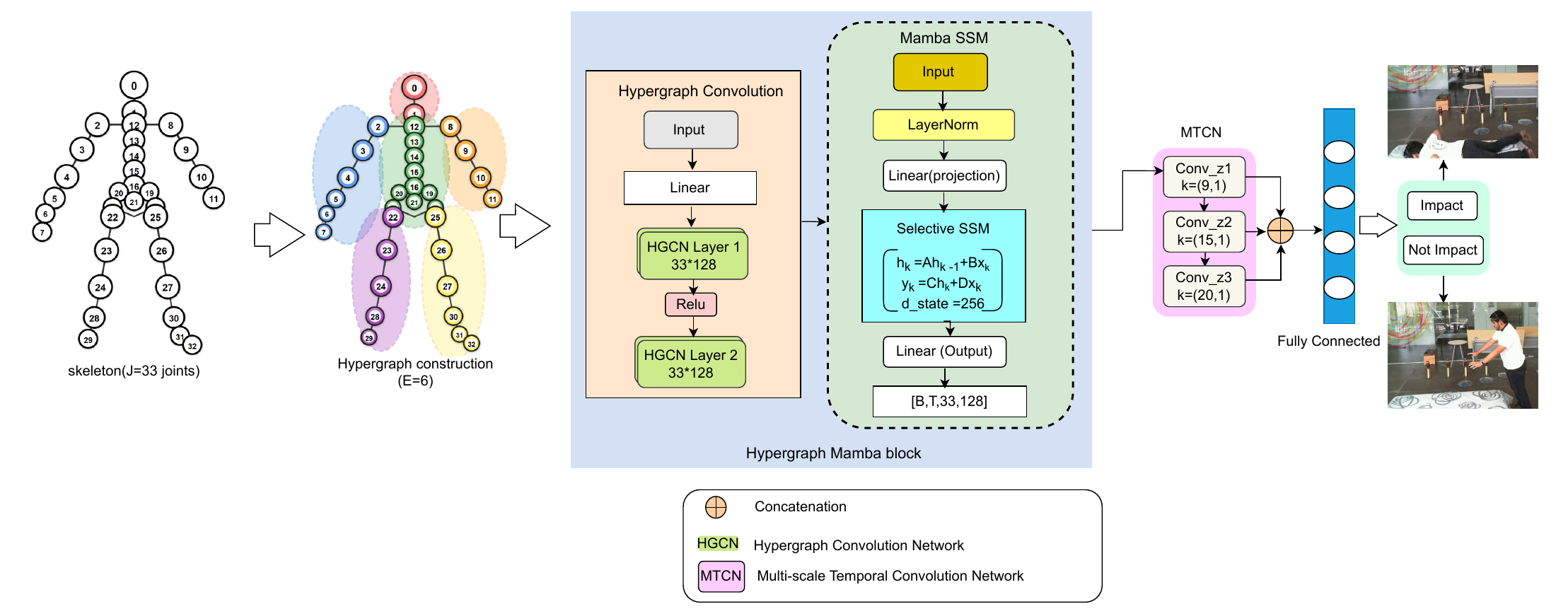}
\caption{Overview of the proposed FLASH architecture. The pipeline consists of four stages: 
\textbf{\ding{172} Input:} 3D skeleton sequence with 33 joints per frame. 
\textbf{\ding{173} Hypergraph Convolution (HGCN):} Two layers using a single-matrix hypergraph with 6 biomechanical hyperedges (torso, left/right arms, left/right legs, head) to encode multi-joint spatial coordination. 
\textbf{\ding{174} Mamba Block:} Selective state-space modeling with hidden dimension 256 to capture long-range temporal dependencies in $O(T)$ time. 
\textbf{\ding{175} Multi-scale Temporal Convolution Network (MTCN):} Parallel convolutions with kernels (9,15,20) followed by a Fully Connected layer for frame-wise impact classification.}
\label{architecture}
\end{figure*}

The complete FLASH pipeline is illustrated in Figure~\ref{architecture}.

\section{Experiments and Results}
\label{sec:experiments}

\subsection{Experimental Setup}

\subsubsection{Datasets}

We evaluate on two skeleton-based fall detection datasets. The 3D Skeletons UP-Fall dataset \cite{koffi2025improved}, built upon the original UP-Fall dataset \cite{martinez2019upfall}, provides 3D skeleton data specifically tailored for impact detection and is used for training and primary evaluation. For cross-dataset evaluation, we test on UMAFall \cite{casilari2017umafall}, which contains similar fall types, enabling direct assessment of generalization capability.
\subsubsection{Evaluation and Implementation Details}
We implement FLASH in PyTorch on an NVIDIA GeForce RTX 5060 GPU (8 GB VRAM), with 80/10/10\% train/val/test splits. The hypergraph uses $J=33$ joints and $E=6$ hyperedges. The 2-layer HGCN has hidden dimension 128, Mamba state dimension 256, and MTCN kernel sizes $\{9, 15, 20\}$. Training uses AdamW ($lr=10^{-4}$, weight decay $10^{-5}$, batch 32, 300 epochs). Performance is assessed via accuracy, precision, recall, specificity, F1-score, FLOPs, and inference time.



\subsection{Comparison with State-of-the-Art}
Table~\ref{tab:sota} compares FLASH with recent state-of-the-art methods on the UP-Fall dataset. Traditional graph-based approaches (ST-GCN \cite{keskes2021vision}, 2s-AGCN \cite{shi2019two}) rely on pairwise joint connections and support only sequence-level classification without frame-level impact localization. Hyper-GNN \cite{hao2021hypergraph} captures higher-order joint relationships but achieves lower accuracy. Transformer-based methods \cite{zhou2022hypergraph} improve temporal modeling but incur quadratic complexity, limiting real-time deployment.
Recently, DistillH-Mamba \cite{koffi2025distillh} proposed a dual-
representation hypergraph approach that processes both first-order (H) and second-order (H²) incidence matrices in parallel pathways, achieving 97.38\% accuracy through knowledge distillation. While this dual-pathway design captures richer structural information, it introduces computational redundancy. In contrast, our single-matrix formulation achieves competitive performance (95.13\% accuracy) with significantly reduced complexity
\begin{table}[t]
\caption{Comparison with State-of-the-Art Methods on UP-Fall Dataset}
\label{tab:sota}
\centering
\adjustbox{max width=\columnwidth}{
\begin{tabular}{lcccc}
\toprule
\textbf{Method} & \textbf{Hypergraph} & \textbf{Acc (\%)} & \textbf{F1 (\%)} & \textbf{Impact} \\
\midrule
ST-GCN \cite{keskes2021vision} & No & 92.16 & 91.31 & No \\
2s-AGCN \cite{shi2019two} & No & 95.10 & -- & No \\
Hyper-GNN \cite{hao2021hypergraph} & Yes (adaptive) & 89.50 & 95.19 & No \\
Transformer \cite{zhou2022hypergraph} & No & 93.15 & 93.80 & No \\
DistillH-Mamba \cite{koffi2025distillh} & Yes (dual + KD) & 97.38 & 97.51 & Yes \\
\textbf{FLASH (Ours)} & \textbf{Yes (single)} & \textbf{95.13} & \textbf{95.52} & \textbf{Yes} \\
\bottomrule
\end{tabular}
}
\end{table}
 that avoids the computational overhead of dual-pathway processing \cite{koffi2025distillh}. Thus, our approach offers an optimal balance between accuracy and computational efficiency for resource constraints devices.

\subsection{Cross-Dataset Generalization on UMAFall}
We evaluate FLASH's generalization through two protocols on the UMAFall 
dataset: (1) \textit{zero-shot transfer}, training on UP-Fall and testing 
directly on UMAFall without fine-tuning, and (2) \textit{in-domain evaluation}, 
training and testing on UMAFall. As shown in Table~\ref{tab:cross_dataset}, 
FLASH achieves strong performance under both settings.
\begin{table}[t]
\caption{Cross-Dataset Zero-Shot Transfer}
\label{tab:cross_dataset}
\centering
\small
\resizebox{\columnwidth}{!}{%
\begin{tabular}{llccccc}
\toprule
\textbf{Train} & \textbf{Test} & \textbf{Acc} & \textbf{Prec} & \textbf{Rec} & \textbf{F1} & \textbf{AUC} \\
\textbf{Dataset} & \textbf{Dataset} & \textbf{(\%)} & \textbf{(\%)} & \textbf{(\%)} & \textbf{(\%)} & \textbf{(\%)} \\
\midrule
UP-Fall & UMAFall (zero-shot) & 95.83 & 93.70 & 98.73 & 96.15 & 98.98 \\
UMAFall & UMAFall (in-domain) & 97.92 & 96.58 & 98.85 & 97.70 & 99.15 \\
\bottomrule
\end{tabular}%
}
\end{table}
In the zero-shot setting, FLASH achieves 95.83\% accuracy on UMAFall without any exposure to this dataset during training. When trained directly on UMAFall, performance improves to 97.92\% accuracy. The gap between zero-shot (95.83\%) and in-domain (97.92\%) evaluation demonstrates that FLASH learns transferable fall representations rather than dataset-specific patterns.
This generalization stems from the biomechanically-grounded hyperedges that encode universal joint coordination patterns independent of recording setup, combined with Mamba's selective temporal modeling that captures distinctive impact patterns common across different fall scenarios.

\subsection{Hypergraph Structure Analysis}

\begin{figure}[t]
\centering
\includegraphics[width=0.95\linewidth]{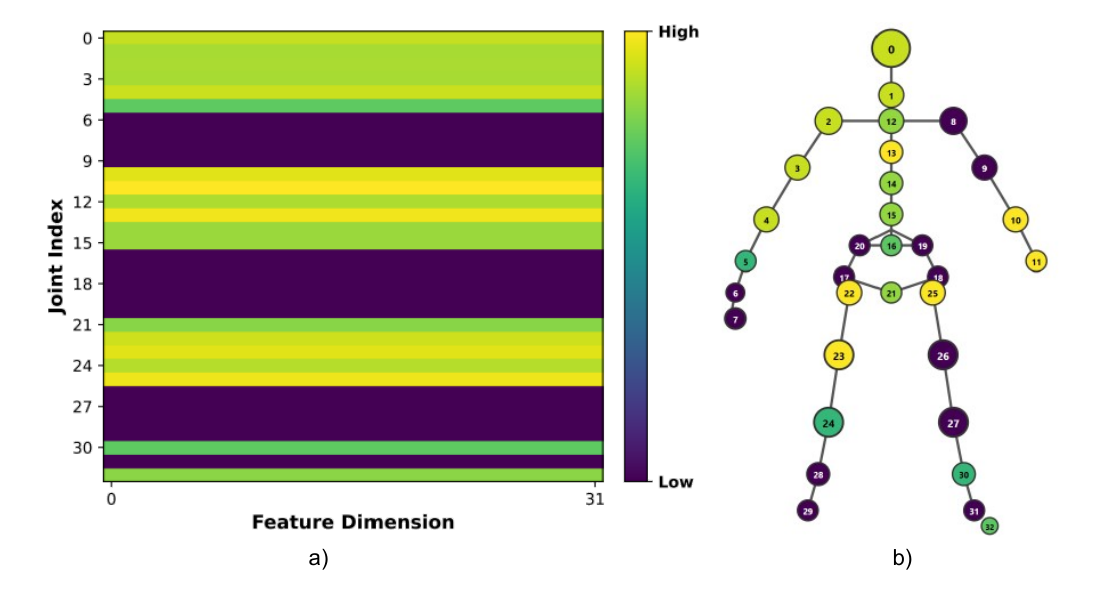}
\caption{Learned joint importance for impact detection.(a) Heatmap showing attention intensity across feature dimensions for each joint. Yellow indicates high and purple the low attention. (b) Skeletal mapping with corresponding attention levels.}
\label{fig:joint_importance}
\end{figure}

To validate that our hypergraph design guides meaningful feature learning, we analyze joint importance patterns during impact detection. Figure~\ref{fig:joint_importance} visualizes attention scores computed from the second hypergraph convolution layer across correctly detected impact frames.
The attention distribution aligns with biomechanical principles. High attention (yellow) concentrates on the torso, knees, and hips primary force transmission hubs and impact points during falls. Medium attention (yellow-green) appears on the head and upper arms, reflecting body orientation tracking and protective arm extension reflexes. Peripheral joints such as fingers and feet receive minimal attention (purple), indicating the model learned to suppress biomechanically non-contributory features. This selective focus validates that hypergraph connectivity enables discriminative feature.

\subsection{Hyperedge Ablation Study}
\label{sec:hyperedge_ablation}

To validate the contribution of each biomechanical hyperedge category, 
we performed systematic ablation by removing each group individually 
while keeping others intact, then evaluating test set accuracy.
\begin{table}[t]
\centering
\caption{Hyperedge ablation study on UP-Fall test set. L+R indicates both left and right hyperedges were removed simultaneously.}

\label{tab:hyperedge_ablation}
\small
\begin{tabular}{lcc}
\toprule
\textbf{Configuration} & \textbf{Accuracy (\%)} & \textbf{Drop (pp)} \\
\midrule
Full Model (6 hyperedges) & 95.13 & — \\
Remove Legs (L+R) & 92.15 & 2.98 \\
Remove Torso & 93.02 & 2.11 \\
Remove Arms (L+R) & 93.80 & 1.33 \\
\bottomrule
\end{tabular}
\end{table}
Results confirm that leg hyperedges are most critical, with their removal 
causing a 2.98pp accuracy drop (95.13\% $\rightarrow$ 92.15\%). This 
validates their role in modeling primary impact sites where the body 
contacts the ground during lateral and forward-knee falls. Torso 
hyperedges show moderate impact (2.11pp drop), reflecting their function 
as the central hub for force transmission from lower to upper body 
segments. Arm hyperedges exhibit the smallest impact (1.33pp drop), 
primarily affecting forward-hand falls where protective extension 
reflexes occur.
The consistent performance degradation across all removals confirms that 
each of the six biomechanical hyperedges contributes uniquely to impact 
detection. Notably, even the least critical category (arms, 1.33pp drop) 
still causes measurable degradation, validating that our design is 
grounded in genuine biomechanical principles rather than arbitrary joint 
groupings.

\subsection{Ablation Study and Computational Cost Analysis}

Table~\ref{tab:ablation_efficiency} compares FLASH against dual-representation 
and transformer-based baselines across spatial and temporal dimensions.
\begin{table}[t]
\caption{Ablation Study and Computational Efficiency Analysis}
\label{tab:ablation_efficiency}
\centering
\adjustbox{max width=\columnwidth}{
\begin{tabular}{lcccc}
\toprule
\textbf{Model} & \textbf{Memory (MB)} & \textbf{Params (M)} & 
\textbf{FLOPs ($\times 10^{7}$)} & \textbf{Time (ms)} \\
\midrule
\multicolumn{5}{l}{\textit{Spatial Representation Comparison}} \\
\quad Hypergraph (Dual + KD) \cite{koffi2025distillh} & 280.46 & 70.12 & 2.266 & 182.3 \\
\quad \textbf{FLASH (Ours)} & \textbf{148.4} & \textbf{37.1} & \textbf{0.65} & \textbf{11.2} \\
\midrule
\multicolumn{5}{l}{\textit{Temporal Modeling Comparison}} \\
\quad Hypergraph (Dual + KD) \cite{koffi2025distillh} & 280.46 & 70.12 & 2.266 & 182.3 \\
\quad Transformer \cite{zhou2022hypergraph} & 341.6 & 85.4 & 4.523 & 245.6 \\
\quad \textbf{FLASH (Ours)} & \textbf{148.4} & \textbf{37.1} & \textbf{0.65} & \textbf{11.2} \\
\bottomrule
\end{tabular}
}
\end{table}
FLASH achieves 47.1\% memory reduction, 71.3\% fewer FLOPs, and 93.9\% faster 
inference vs. the dual hypergraph baseline, and 56.6\% less memory, 85.6\% 
fewer FLOPs, and 95.4\% faster inference vs. the Transformer. With 148.4 MB 
and 11.2 ms inference time, FLASH is suitable for edge deployment on smartphones, Raspberry Pi, and NVIDIA Jetson platforms.

\subsection{Robustness to Occlusion}
\label{sec:robustness}

To evaluate robustness under real-world occlusion, we performed random joint 
dropout at rates of 10\%, 20\%, and 30\% on the test set. As shown in 
Table~\ref{tab:robustness}, FLASH maintains 93.75\% accuracy with 30\% of 
joints missing (only 1.38pp drop from baseline), owing to hyperedges spanning 
multiple joints that provide redundant biomechanical pathways when individual 
joints are occluded. This validates FLASH's deployability in healthcare 
environments where camera occlusion is common.

\begin{table}[t]
\centering
\caption{Performance under partial occlusion.}
\label{tab:robustness}
\small
\begin{tabular}{lcc}
\toprule
\textbf{Dropout Rate} & \textbf{Accuracy (\%)} & \textbf{Drop (pp)} \\
\midrule
0\% (Baseline) & 95.13 & — \\
10\% & 94.84 & 0.29 \\
20\% & 94.50 & 0.63 \\
30\% & 93.75 & 1.38 \\
\bottomrule
\end{tabular}
\end{table}

\section{Conclusion}
\label{sec:conclusion}

We presented FLASH, integrating single-matrix hypergraphs with Mamba's selective 
state-space models for impact fall detection. Biomechanically-grounded hyperedges 
capture multi-joint coordination while Mamba's linear-time complexity ensures 
real-time efficiency. FLASH achieves 95.13\% accuracy and 95.52\% F1 on UP-Fall, 
with 95.83\% zero-shot transfer to UMAFall, 37.1M parameters, and 11.2 ms 
inference. Future work will address real-world validation, 
adaptive hyperedge learning, and extension to multi-person scenarios.
\bibliographystyle{IEEEbib}
\bibliography{refs}

\end{document}